\documentclass[11pt,a4paper]{article}
\usepackage[hyperref]{acl2021}
\usepackage{times}
\usepackage{latexsym}

\usepackage{subfigure}
\usepackage{microtype}

\aclfinalcopy 
\def\aclpaperid{1053} 

\usepackage[T1]{fontenc}
\usepackage[utf8]{inputenc}
\usepackage{floatrow}

\usepackage{comment}
\usepackage{hyperref}       
\usepackage{url}            
\usepackage{booktabs}       
\usepackage{amsfonts}       
\usepackage{nicefrac}       
\usepackage{microtype}      
\usepackage{graphics,graphicx}
\usepackage{wrapfig}
\usepackage{enumitem}
\usepackage{color}
\usepackage{tikz}
\usepackage{tipa}
\usepackage{arydshln}
\usepackage{CJKutf8}
\usepackage{multirow,multicol}

\usepackage{bm,bbm,dsfont}
\usepackage{amsmath, amssymb}
\usepackage{pgfplots}
\usepackage{enumitem}

\usepackage{amsmath}
\usepackage{tabularx}
\usepackage{booktabs}
\usepackage{amssymb}
\usepackage{pifont}
\usepackage{bm}
\usepackage{multirow}
\usepackage{tcolorbox}
\usepackage[framemethod=tikz]{mdframed}
\usepackage{tikz,pgfplots}
\usepgfplotslibrary{groupplots}
\pgfplotsset{compat=1.13}

\definecolor{g-red}{HTML}{DB4437}
\definecolor{g-blue}{HTML}{4285F4}
\definecolor{g-green}{HTML}{0F9D58}
\definecolor{g-yellow}{HTML}{F4B400}
\definecolor{g-orange}{HTML}{FF9800}
\definecolor{g-grey}{HTML}{9E9E9E}
\definecolor{shannon}{HTML}{304FFE}
\definecolor{uw}{RGB}{138,43,226}
\definecolor{stanford}{RGB}{255,69,0}
\definecolor{const}{RGB}{68, 110, 182}
\definecolor{head}{RGB}{246, 180, 32}
\definecolor{freq}{RGB}{0, 0, 0}

\newmdenv[innerlinewidth=0.5pt, roundcorner=4pt,linecolor=black,innerleftmargin=6pt,
innerrightmargin=6pt,innertopmargin=6pt,innerbottommargin=6pt]{examplebox}

\title{ChineseBERT: Chinese Pretraining Enhanced by\\ Glyph and Pinyin Information}
\author{
Zijun Sun$^\clubsuit$, Xiaoya Li$^\clubsuit$, Xiaofei Sun$^\clubsuit$, Yuxian Meng$^\clubsuit$, Guoyin Wang$^\blacktriangledown$\\
{\bf Xiang Ao$^\spadesuit$, Qing He$^\spadesuit$, Fei Wu$^\blacklozenge$ and Jiwei Li$^{\blacklozenge\clubsuit}$}\\
   $^\clubsuit$Shannon.AI, 
  $^\blacklozenge$Zhejiang University, $^\blacktriangledown$Amazon\\
  $^\spadesuit$Key Lab of Intelligent Information Processing of Chinese Academy of Sciences  \\
  \{zijun\_sun, xiaoya\_li, xiaofei\_sun, yuxian\_meng, jiwei\_li\}@shannonai.com\\
  \{aoxiang,heqing\}@ict.ac.cn,
  wufei@zju.edu.cn
}

\begin{document}
\begin{CJK*}{UTF8}{gbsn}

\maketitle

\begin{abstract}
Recent pretraining models in Chinese neglect  two important aspects  specific to the Chinese language: glyph and pinyin,  which carry significant syntax and semantic  information for language understanding.  In this work, we propose ChineseBERT, which  incorporates both the {\it glyph} and {\it pinyin} information of Chinese characters into language model pretraining. The glyph embedding is obtained   based on different fonts of a Chinese character, being able to capture character semantics from the visual features, and the pinyin embedding  characterizes the pronunciation of Chinese characters, which    handles the highly prevalent heteronym phenomenon in Chinese (the same character has different pronunciations with different meanings).   Pretrained on large-scale unlabeled Chinese corpus, the proposed ChineseBERT model yields significant performance boost over baseline models with fewer training steps. The proposed model achieves new SOTA performances on  a wide range of Chinese NLP tasks， including machine reading comprehension, natural language inference, text classification, sentence pair matching, and competitive performances in   named entity recognition and word segmentation.\footnote{The code and pretrained models are publicly available at \url{https://github.com/ShannonAI/ChineseBert}.}
\footnote{To appear at ACL2021.}
\end{abstract}

\section{Introduction}

Large-scale pretrained models have become a fundamental backbone for various natural language processing tasks such as natural language understanding \citep{yinhan2019roberta}, text classification \citep{reimers2019sentence,chai2020description} and question answering \citep{clark2017simple,lewis2020retrieval}. Apart from English NLP tasks, pretrained models have also demonstrated their effectiveness for various Chinese NLP tasks \citep{sun2019ernie,sun2020ERNIE,cui2019pre,cui2020revisiting}.

Since   pretraining models are originally designed for English,  two important aspects  specific to the Chinese language are
 missing 
 in current large-scale pretraining: glyph-based information and pinyin-based information. 
For the former, a key aspect  
 that makes Chinese distinguishable from  languages such as English, German, is that Chinese is a logographic language. 
 The logographic of characters 
 encodes semantic information. For example, ``液 (liquid)'', ``河 (river)'' and ``湖 (lake)'' all have the radical ``氵(water)'', which indicates that they are all related to water in semantics. 
Intuitively, the rich semantics behind Chinese character glyphs should enhance the expressiveness of Chinese NLP models. This idea has motivated a variety of of work on learning and incorporating Chinese glyph information into neural models \citep{sun2014radical,shi2015radical,liu2017learning,dai2017glyph,su2017learning,meng2019glyce}, but not yet large-scale pretraining.

For the latter, {\it pinyin}, the Romanized sequence of a Chinese character representing its pronunciation(s), is  crucial in 
modeling 
both 
semantic and syntax 
information that can not be captured by contextualized or glyph embeddings.
This  aspect is especially important considering the highly prevalent heteronym phenomenon in Chinese\footnote{Among 7000 common characters in Chinese, there are about 700 characters that have multiple pronunciations, according to the Contemporary Chinese Dictionary.}, where 
the same character have multiple pronunciations, each of which is associated with a specific meaning. 
Each pronunciation is associated with a specific pinyin expression. 
At the semantic level, for example, the Chinese character ``乐'' has two distinctly different pronunciations:
``乐'' can be pronounced as 
 ``yu\`e [y\textepsilon$^{51}$]'', which means ``music'', and ``l\`e [l\textipa{G}$^{51}$]'', which means ``happy''. 
On the syntax level, pronunciations help identify the part-of-speech of a character. 
  For example,  character ``还'' has two pronunciations: ``hu\'{a}n[xwan$^{35}$]'' and ``h\'{a}i[xa\textsci$^{35}$]'', with the former meaning the verb ``return'' and the latter meaning the adverb ``also''.
Different pronunciations of the same character cannot be distinguished by 
  the glyph embedding since the logographic is the same, or  the char-ID embedding, since they both point to the same character ID, but can be characterized by pinyin. 

In this work, we propose ChineseBERT, a model that incorporates the glyph and pinyin information of Chinese characters into  the process of large-scale pretraining. The glyph embedding is based on different fonts of a Chinese character, being able to capture character semantics from the visual surface character forms. The pinyin embedding models different semantic meanings that share the same character form and thus bypasses the limitation of interwound morphemes behind a single character. For a Chinese character, the glyph embedding, the pinyin embedding and the character embedding are combined to form a fusion embedding, which models the distinctive semantic property of that character.

With less training data and fewer training epochs, ChineseBERT achieves significant performance boost 
over baselines
across a wide range of Chinese NLP tasks.
It achieves new SOTA performances on 
 a wide range of Chinese NLP tasks，
 including machine reading comprehension, natural language inference, text classification, sentence pair matching,
 and results comparable to SOTA performances in 
  named entity recognition and word segmentation. 


\section{Related Work}
\subsection{Large-Scale Pretraining in NLP}
Recent years has witnessed substantial work on large-scale pretraining in NLP. BERT \citep{devlin2018bert}, which is built on top of the Transformer architecture \citep{vaswani2017attention}, is pretrained on large-scale unlabeled text corpus in the manner of Masked Language Model (MLM) and Next Sentence Prediction (NSP). Following this trend, considerable progress has been made by modifying the masking strategy \citep{yang2019xlnet,danqi2020spanbert}, pretraining tasks \citep{liu-etal-2019-multi,Clark2020ELECTRA} or model backbones \citep{Lan2020ALBERT,lample2019large,performer}. Specifically, RoBERTa \citep{yinhan2019roberta} proposed to remove the NSP pretraining task since it has been proved to offer no benefits for improving downstream performances. The GPT series \citep{radford2019gpt2,brown2020language} and other BERT variants \citep{lewis2019bart,song2019mass,lample2019cross,NEURIPS2019_c20bb2d9,bao2020unilmv2,Zhu2020Incorporating} adapted the paradigm of large-scale unsupervised pretraining to text generation tasks such as machine translation, text summarization and dialog generation, so that generative models can enjoy the benefit of large-scale pretraining.

Unlike the English language, Chinese has its particular characteristics in terms of syntax, lexicon and pronunciation. Hence, pretraining Chinese models should fit the Chinese features correspondingly. \citet{li-etal-2019-word-segmentation} proposed to use Chinese character as the basic unit instead of word or sub-word that is used in English \citep{wu2016googles,sennrich-etal-2016-neural}. ERNIE \citep{sun2019ernie,sun2020ERNIE} applied three types of masking strategies -- char-level masking, phrase-level masking and entity-level masking -- to enhance the ability of capturing multi-granularity semantics. \citet{cui2019pre,cui2020revisiting} pretrained models using the Whole Word Masking strategy, where all characters within a Chinese word are masked altogether. In this way, the model is learning to address a more challenging task as opposed to predicting word components.
More recently, \citet{zhang2020cpm} developed the largest Chinese pretrained language model to date -- CPM. It is pretrained on 100GB Chinese data and has 2.6B parameters comparable to ``GPT3 2.7B'' \citep{brown2020language}.
\citet{xu-etal-2020-clue} released the first large-scale Chinese Language Understanding Evaluation benchmark CLUE, facilitating researches in large-scale Chinese pretraining.

\subsection{Learning Glyph Information}
Learning glyph information from surface Chinese character forms has gained attractions since the prevalence of deep neural networks.
Inspired by word embeddings \citep{mikolov2013distributed,mikolov2013efficient}, \citet{sun2014radical,shi2015radical,li2015component,yin2016multi} used indexed radical embeddings to capture character semantics, improving model performances on a wide range of Chinese NLP tasks.
Another way of incorporating glyph information is to view characters in the form of image, by which glyph information can be naturally learned through image modeling. However, early work on learning visual features  is not smooth. \citet{liu2017learning,shao2017character,zhang2017encoding,dai2017glyph} used CNNs to extract glyph features from character images but did not achieve consistent performance boost over all tasks.
\citet{su2017learning,tao2019chinese} obtained positive results on the word analogy and word similarity tasks but they did not further evaluate the learned glyph embeddings on more tasks. \citet{meng2019glyce}  applied glyph embeddings to a broad array of Chinese tasks. They designed a specific CNN structure 
for character feature extraction
and used image classification as an auxiliary objective to regularize the influence of a limited number of images. \citet{Song_Sehanobish_2020,xuan2020fgn} extended the idea of \citet{meng2019glyce} to the task of named entity recognition (NER), significantly improving performances against vanilla BERT models.

\section{Model}
\begin{figure}
    \centering
    \includegraphics[scale=0.55]{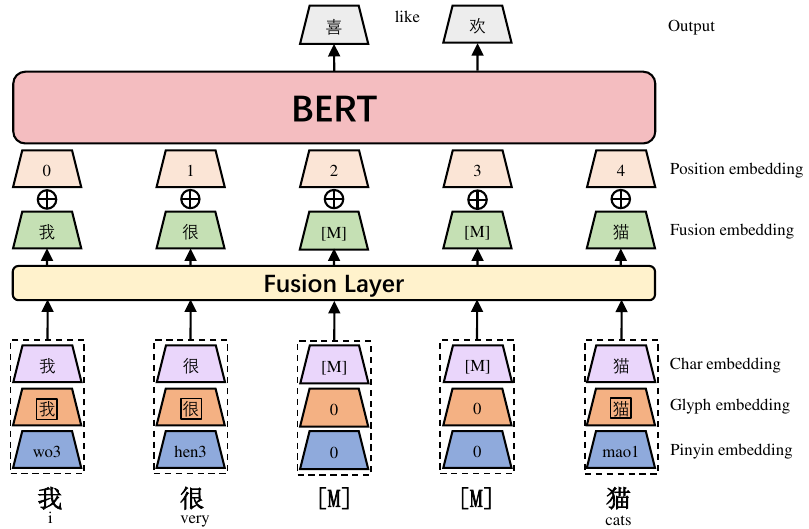}
    \caption{An overview of  ChineseBERT. The fusion layer consumes three D-dimensional embeddings -- char embedding, glyph embedding and pinyin embedding.
    The three  embeddings are first concatenated, and then mapped to a D-dimensional embedding through a fully connected layer to form the fusion embedding.
    }
    \label{fig:overview}
\end{figure}

\begin{figure}[!ht]
  \centering
  \includegraphics[scale=0.7]{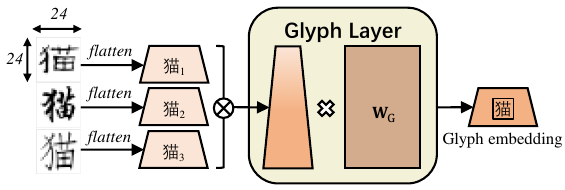}
  \caption{An overview of inducing the {\it glyph} embedding. $\bigotimes$ denotes vector concatenation. For each Chinese character, we use three types of fonts -- {\it FangSong}, {\it XingKai} and {\it LiShu}, each of which is a $24\times 24$ image with pixel value ranging $0\sim 255$.  Images are concatenated into a tensor of size $24\times 24\times 3$. The tensor is flattened and 
   passed to an FC to obtain  the glyph embedding.}
  \label{fig:glyph}
\end{figure}

\begin{figure}[!ht]
  \centering
  \includegraphics[scale=0.8]{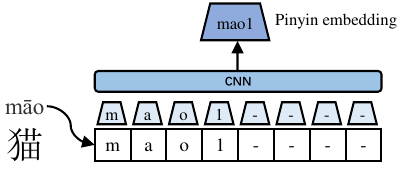}
  \caption{An overview of inducing the {\it pinyin} embedding. For any Chinese character, e.g. 猫 (cat) in this case, a CNN with width 2 is applied to the sequence of Romanized pinyin letters, followed by max-pooling to derive the final pinyin embedding.}
  \label{fig:pinyin}
\end{figure}

\begin{figure}[!ht]
  \centering
  \includegraphics[scale=0.8]{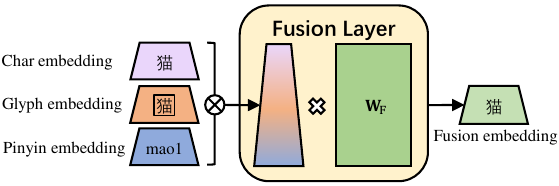}
  \caption{An overview of the fusion layer. $\bigotimes$ denotes vector concatenation, and $\times$ is vector-matrix multiplication. We  concatenate the char embedding, the glyph embedding and the pinyin embedding, and use an FC layer with a learnable matrix $\mathbf{W}_F$ to induce the fusion embedding.}
  \label{fig:fusion}
\end{figure}

\subsection{Overview}
Figure \ref{fig:overview} shows an overview of the proposed ChineseBERT model. For each Chinese character, its char embedding, glyph embedding and pinyin embedding are first concatenated, and then mapped to a D-dimensional embedding through a fully connected layer to form the fusion embedding.
    The fusion embedding is then added with the position embedding, which is fed  as input to the BERT model
Since we do not use the NSP pretraining task, we omit the segment embedding.
We use both Whole Word Masking (WWM) \citep{cui2019pre} and Char Masking (CM) for pretraining (See Section \ref{sec:masking} for details).

\subsection{Input}
The input to the model is the addition of the learnable absolute positional embedding and the fusion embedding, where the fusion embedding is based on the char embedding, the glyph embedding and the pinyin embedding of the corresponding character. The char embedding performs in a way analogous to the token embedding used in BERT but at the character granularity.
Below we respectively describe how to induce the glyph embedding, the pinyin embedding and the fusion embedding.

\paragraph{Glyph Embedding}
We followed \citet{meng2019glyce} to use three types of Chinese fonts -- FangSong, XingKai and LiShu, each of which is instantiated as a $24\times 24$ image with floating point pixels ranging from 0 to 255. 
Different from \citet{meng2019glyce}, which used CNNs to convert image to representations, 
we use an FC layer. 
We 
 first converted 
the 24$\times $24$\times$3 vector to a 2,352 vector.
The flattened vector is 
 fed to an FC layer to obtain the output glyph vector. 

\paragraph{Pinyin Embedding}
The {\it pinyin} embedding for each character is used to decouple different semantic meanings belonging to the same character form, as shown in Figure \ref{fig:pinyin}.
We use the opensourced pypinyin package\footnote{\url{https://pypi.org/project/pypinyin/}} to generate  pinyin sequences for its constituent characters. 
pypinyin is a  system that combines machine learning models with 
dictionary-based rules  to infer the pinyin for characters given contexts. 
Pinyin for a Chinese character is  a sequence of 
 Romanian characters, with one of four diacritics denoting tones.
 We use special tokens to denote tones, which are appended to the end of the Romanian character sequence. 
We apply a  CNN  model with width 2 on the pinyin sequence, followed by max-pooling to derive the resulting pinyin embedding. 
This makes output dimensionality immune to the length of the input pinyin sequence. 
The length of the input pinyin sequence is fixed at 8, with the remaining slots filled with a special letter ``-'' when the actual length of the pinyin sequence does not reach 8.

\paragraph{Fusion Embedding}
Once we have the char embedding, the glyph embedding and the pinyin embedding for a character, we concatenate them to form a $3D$-dimensional vector. 
The fusion layers maps the $3D$-dimensional vector to $D$-dimensional through a fully connected layer. 
The fusion embedding is added with position embedding, and output to the BERT layer. 
An illustration is shown in Figure \ref{fig:fusion}.

\subsection{Output}
The output  is the corresponding contextualized representation for each input Chinese character  \citep{devlin2018bert}.  

\section{Pretraining Setup}
\subsection{Data}
We collected our pretraining data from CommonCrawl\footnote{\url{https://commoncrawl.org/}}. After pre-processing (such as removing the data with too much English text and filtering the \texttt{html} tagger), about 10\% high-quality data is maintained for pretraining, containing  4B Chinese characters in total. We use the LTP toolkit\footnote{\url{http://ltp.ai/}} \citep{che-etal-2010-ltp} to identify the boundary of Chinese words for whole word masking. 

\subsection{Masking Strategies}
\label{sec:masking}
We use two masking strategies -- Whole Word Masking (WWM) and Char Masking (CM) for ChineseBERT. \citet{li-etal-2019-word-segmentation} suggested that using Chinese characters as the basic input unit can alleviate the out-of-vocabulary issue in the Chinese language. We thus adopt the method of masking random characters in the given context, denoted by Char Masking.
On the other hand, a large number of words in Chinese consist of multiple characters, for which the CM strategy may be too easy for the model to predict. 
For example, for the input context ``我喜欢逛紫禁[M] (i like going to The Forbidden [M])'', the model can easily predict that the masked character is ``城 (City)''.
Hence, we follow \citet{cui2019pre} to use WWM, a strategy to mask out all characters within a selected word, mitigating the easy-predicting shortcoming of the CM strategy.
Note that for both WWM and CM, the basic input unit is Chinese characters. The main difference between WWM and CM lies in how they mask characters and how the model predicts masked characters.

\subsection{Pretraining Details}
Different from \citet{cui2019pre} who pretrained their model based on the official pretrained Chinese BERT model, we train the ChineseBERT model from scratch. 
To enforce the model to learn both long-term and short-term dependencies, we propose to alternate pretraining between {\it packed input} and {\it single input}, where the packed input is the concatenation of multiple sentences with a maximum length 512, and the single input is a single sentence.
We feed the packed input with probability of 0.9 and the single input with probability of 0.1.
We apply Whole Word Masking 90\% of the time and Char Masking 10\% of the time. The masking probability for each word/char is 15\%. If the $i$-th word/char is chosen, we mask it 80\% of
the time, replace it with a random word/char 10\% of the time and maintain it 10\% of the time.
We also use the dynamic masking strategy to avoid duplicate training instances \citep{yinhan2019roberta}.
We use two model setups: \texttt{base} and \texttt{large}, respectively consisting of 12/24 Transformer layers, with input dimensionality of 768/1,024 and 12/16 heads per layer.
This makes our models comparable to other BERT-style models in terms of model size. 
Upon the submission of the paper, we have trained the \texttt{base} model 500K steps with a maximum learning rate 1e-4, warmup of 20K steps  and a batch size of 3.2k,
and the \texttt{large} model 280K steps with   a maximum learning rate 3e-4, warmup of 90K steps and a batch size of 8k. 
After pretraining, the model can be directly used to be finetuned on downstream tasks in the same way as BERT \citep{devlin2018bert}.


\begin{table}[t]
    \centering
    \small
    \scalebox{0.6}{
    \begin{tabular}{lcccc}
    \toprule
    & {\bf ERNIE} & {\bf BERT-wwm} & {\bf MacBERT} & {\bf ChineseBERT} \\\midrule
    Data Source & Heterogeneous & Wikipedia & Heterogeneous & CommonCrawl \\
    Vocab Size & 18K & 21K & 21K & 21K\\
    Input Unit & Char & Char &  Char & Char\\
    Masking & T/P/E & WWM & WWM/N & WWM/CM \\
    Task & MLM/NSP & MLM & MAC/SOP & MLM\\
    Training Steps & - & 2M & 1M & 1M \\
    Init Checkpoint &   & BERT & BERT & random\\
    \# Token & -- & 0.4B & 5.4B & 5B \\
    \bottomrule
    \end{tabular}
    }
    \caption{Comparison of data statistics between ERNIE \citep{sun2019ernie}, BERT-wwm \citep{cui2019pre}, MacBERT \citep{cui2020revisiting} and our proposed ChineseBERT. T: Token, P: Phrase, E: Entity, WWM: Whole Word Masking, N: N-gram, CM: Char Masking, MLM: Masked Language Model, NSP: Next Sentence Prediction, MAC: MLM-As-Correlation. SOP: Sentence Order Prediction.}
    \label{tab:statistics}
\end{table}

\section{Experiments}
We conduct extensive experiments on a variety of Chinese NLP tasks. Models are separately finetuned on task-specific datasets for evaluation. Concretely, we use the following tasks:
\begin{itemize}
  \item { Machine Reading Comprehension (MRC)}
  \item { Natural Language Inference (NLI)}
  \item { Text Classification (TC)}
    \item { Sentence Pair Matching (SPM)}
  \item { Named Entity Recognition (NER)} 
  \item { Chinese Word Segmentation (CWS)}. 
\end{itemize}

We compare ChineseBERT to current state-of-the-art ERNIE \citep{sun2019ernie,sun2020ERNIE}, BERT-wwm \citep{cui2019pre} and MacBERT \citep{cui2020revisiting} models. ERNIE adopts various masking strategies including token-level, phrase-level and entity-level masking to pretrain BERT on large-scale heterogeneous data. BERT-wwm/RoBERTa-wwm continues pretraining on top of official pretrained Chinese BERT/RoBERTa models with the Whole Word Masking pretraining strategy. 
Unless otherwise specified, we use BERT/RoBERTa to represent BERT-wwm/RoBERTa-wwm and omit ``wwm''.
MacBERT improves upon RoBERTa by using the MLM-As-Correlation (MAC) pretraining strategy as well as the sentence-order prediction (SOP) task. 
It is worth noting that BERT and BERT-wwm do not have the large version available online, and we thus omit the  corresponding performances. 

A comparison of these models is shown in Table \ref{tab:statistics}.
It is {\bf worth noting} that  training steps  
of the proposed model
 significantly smaller than baseline models. 
Different from BERT-wwm and MacBERT which are initialized with pretrained BERT, the proposed model is initialized from scratch. Due to the additional consideration of glyph and pinyin, the proposed cannot be directly initialized using a vanilla BERT model, as the model structures are different.
Even initialized from scratch, the proposed model is trained fewer steps than 
the steps in 
retraining BERT-wwm and MacBERT after BERT initialization. 

\subsection{Machine Reading Comprehension}
Machine reading comprehension tests the model's ability of answering the questions based on the given contexts. We use two datasets for this task: CMRC 2018 \citep{cui-emnlp2019-cmrc2018} and CJRC \citep{Duan_2019} .
CMRC is a span-extraction style dataset while CJRC additionally has yes/no questions and no-answer questions.
CMRC 2018 and CJRC respectively contain 10K/3.2K/4.9K and 39K/6K/6K data instances for training/dev/test. Test results for CMRC 2018 are evaluated from the CLUE leaderboard.\footnote{\url{https://github.com/CLUEbenchmark/CLUE}}
Note that the CJRC dataset is different from the one used in \citet{cui2019pre} as \citet{cui2019pre} did not release their train/dev/test split. We thus run the released models on the CJRC dataset used in this work for comparison.

Results are shown in Table \ref{tab:cmrc} and Table \ref{tab:cjrc}.
As we can see, ChineseBERT yields significant performance boost on both datasets, and the improvement of EM is larger than that of F1 on the CJRC dataset, which indicates that ChineseBERT is better at detecting exact answer spans.

\begin{table}[t]
  \centering
  \small
  \begin{tabular}{lcc}
    \toprule 
    &\multicolumn{2}{c}{{\it CMRC}}\\
   {\bf Model} & {\bf Dev} & {\bf Test}\\\midrule
    & \multicolumn{2}{c}{\underline{\texttt{Base}}}\\
    ERNIE & 66.89 & 74.70  \\
    BERT & 66.77 & 71.60 \\
    BERT$^{\circ}$  & 66.96 & 73.95  \\
    RoBERTa$^{\circ}$  &  67.89 & 75.20  \\
    MacBERT & -- & --  \\
    ChineseBERT  &  {\bf 67.95} & {\bf 75.35}    \\
    \cdashline{1-3}
    & \multicolumn{2}{c}{\underline{\texttt{Large}}}\\
    RoBERTa$^{\circ}$  & 70.59 & 77.95 \\
    MacBERT  & -- & --  \\
    ChineseBERT  & {\bf 70.70} &  {\bf 78.05} \\
    \bottomrule
  \end{tabular}
  \caption{Performances of different models on CMRC. EM is reported for comparison. $\circ$ represents models pretrained on extended data.}
  \label{tab:cmrc}
\end{table}

\begin{table}[t]
  \centering
  \small
  \begin{tabular}{lcccc}
    \toprule 
    & \multicolumn{4}{c}{{\it CJRC}}\\
     & \multicolumn{2}{c}{{\bf Dev}} & \multicolumn{2}{c}{{\bf Test}}\\
    {\bf Model} & {\bf EM} & {\bf F1} & {\bf EM} & {\bf F1} \\\midrule
    & \multicolumn{4}{c}{\underline{\texttt{Base}}}\\
    BERT & 59.8 & 73.0 & 60.2 & 73.0 \\
    BERT$^{\circ}$  & 60.8 & 74.0 & 61.4 & 73.9 \\
    RoBERTa$^{\circ}$  & 62.9 & 76.6 & 63.8 & 76.6\\
    ChineseBERT & {\bf 65.2} & {\bf 77.8} & {\bf 66.2} & {\bf 77.9}\\
    \cdashline{1-5}
    & \multicolumn{4}{c}{\underline{\texttt{Large}}}\\
    RoBERTa$^{\circ}$  & 65.6 & 77.5 & 66.4 & 77.6\\
    ChineseBERT & {\bf 66.5} & {\bf 77.9} & {\bf 67.0} & {\bf 78.3} \\
    \bottomrule
  \end{tabular}
  \caption{Performances of different models on the MRC dataset CJRC. We report results for baseline models based on their released models.  $\circ$ represents models pretrained on extended data.}
  \label{tab:cjrc}
\end{table}

\subsection{Natural Language Inference (NLI)}
  The goal of NLI is to determine the entailment relationship between a hypothesis and a premise. We use the Cross-lingual Natural Language Inference (XNLI) dataset \citep{conneau2018xnli} for evaluation.
The  corpus is a crowd-sourced collection of 5K test and 2.5K dev pairs for the MultiNLI corpus. Each sentence pair is annotated with the ``entailment'', ``neutral'' or ``contradiction'' label. We use the official machine translated Chinese data for training.\footnote{\url{https://github.com/facebookresearch/XNLI}}

Results are present in Table \ref{tab:xnli}, which shows that ChineseBERT is able to achieve the best performances for both \texttt{base} and \texttt{large} setups.

\begin{table}[t]
  \centering
  \small
  \begin{tabular}{lcc}
    \toprule 
    &\multicolumn{2}{c}{{\it XNLI}}\\
   {\bf Model} & {\bf Dev} & {\bf Test}\\\midrule
    & \multicolumn{2}{c}{\underline{\texttt{Base}}}\\
    ERNIE & 79.7 & 78.6  \\
    BERT & 79.0 & 78.2 \\
    BERT$^{\circ}$  & 79.4 & 78.7   \\
    RoBERTa$^{\circ}$  & 80.0 & 78.8  \\
    MacBERT  & 80.3 & 79.3  \\
    ChineseBERT  &  {\bf 80.5}&{\bf 79.6}    \\
    \cdashline{1-3}
    & \multicolumn{2}{c}{\underline{\texttt{Large}}}\\
    RoBERTa$^{\circ}$  & 82.1 & 81.2 \\
    MacBERT  & 82.4 & 81.3  \\
    ChineseBERT  & {\bf 82.7} &  {\bf 81.6} \\
    \bottomrule
  \end{tabular}
  \caption{Performances of different models on XNLI. Accuracy is reported for comparison. $\circ$ represents models pretrained on extended data.}
  \label{tab:xnli}
\end{table}

\subsection{Text Classification (TC)}
In text classification
  the model is required to categorize a piece of text into one of the specified classes. We follow \citet{cui2019pre} to use THUCNews \citep{li2007scalable} and ChnSentiCorp\footnote{\url{https://github.com/pengming617/bert_classification/tree/master/data}} for this task.  
     THUCNews is a subset of THUCTC \footnote{\url{http://thuctc.thunlp.org/}}, with 50K/5K/10K data points respectively for training/dev/test. Data is evenly distributed in 10 domains including sports, finance, etc.\footnote{\url{https://github.com/gaussic/text-classification-cnn-rnn}}
ChnSentiCorp is a binary sentiment classification dataset containing 9.6K/1.2K/1.2K data points respectively for training/dev/test. 
  The two datasets are relatively simple with vanilla BERT  achieving an accuracy of above 95\%. Hence, apart from  THUCNews and ChnSentiCorp, we also use TNEWS, a more difficult dataset that is included in the CLUE benchmark \citep{xu-etal-2020-clue}.\footnote{\url{https://github.com/CLUEbenchmark/CLUE}}
 TNEWS is a 15-class short news text classification dataset with 53K/10K/10K data points for training/dev/test. 

Results are shown in Table \ref{tab:tc}.
On ChunSentiCorp and THUCNews, the improvement from ChineseBERT is marginal as baselines have already achieved quite high results on these two datasets. On the TNEWS dataset, ChineseBERT outperforms all other models. We can see that the ERNIE model only performs slightly worse than  ChineseBERT. This is because ERNIE is trained on additional web data, which is beneficial to model web news text that covers a wide range of domains.

\begin{table}[t]
  \centering
  \small
  \scalebox{0.85}{
  \begin{tabular}{lcccccc}
    \toprule 
     & \multicolumn{2}{c}{{\it ChnSentiCorp}} & \multicolumn{2}{c}{{\it THUCNews}}
     & \multicolumn{2}{c}{{\it TNEWS}}\\
    {\bf Model} & {\bf Dev} & {\bf Test} & {\bf Dev} & {\bf Test} & {\bf Dev} & {\bf Test}\\
    \midrule
    & \multicolumn{6}{c}{\underline{\texttt{Base}}}\\
    ERNIE & 95.4 & 95.5 & 97.6 & 97.5&58.24 & 58.33 \\
    BERT & 95.1 & 95.4 & 98.0 & 97.8 & 56.09 & 56.58\\
    BERT$^{\circ}$  &  95.4 & 95.3 & 97.7 & 97.7  & 56.77 & 56.86\\
    RoBERTa$^{\circ}$  & 95.0 & 95.6 & {\bf 98.3} & 97.8  & 57.51 & 56.94\\
    MacBERT  & 95.2 & 95.6 & 98.2 & 97.7 & -- & --\\ 
    ChineseBERT  & {\bf 95.6} & {\bf 95.7} & 98.1 & {\bf 97.9} & {\bf 58.64}  & {\bf 58.95} \\
    \cdashline{1-7}
    & \multicolumn{6}{c}{\underline{\texttt{Large}}}\\
    RoBERTa$^{\circ}$  & {\bf 95.8} & 95.8 & {\bf 98.3} & 97.8 & 58.32 & 58.61\\
    MacBERT & 95.7 & {\bf 95.9} & 98.1 & {\bf 97.9} & -- & --\\
    ChineseBERT & {\bf 95.8} & {\bf 95.9} & {\bf 98.3} & {\bf 97.9} & {\bf 59.06} & {\bf 59.47}  \\
    \bottomrule
  \end{tabular}
  }
  \caption{Performances of different models on TC datasets ChnSentiCorp, THUCNews and TNEWS.  
  The results of TNEWS are taken from the CLUE paper \citep{xu-etal-2020-clue}.
  Accuracy is reported for comparison. $\circ$ represents models pretrained on extended data.}
  \label{tab:tc}
\end{table}

\subsection{Sentence Pair Matching (SPM)}

For SPM, the model is asked to determine whether a given sentence pair expresses the same semantics. We use the LCQMC \citep{liu2018lcqmc} and BQ Corpus  \citep{chen-etal-2018-bq}  datasets for evaluation. LCQMC is a large-scale Chinese question matching corpus for judging whether two given questions have the same intent, with 23.9K/8.8K/12.5K sentence pairs for training/dev/test.
BQ Corpus is another large-scale Chinese dataset containing 100K/10K/10K sentence pairs for training/dev/test.
Results are shown in Table \ref{tab:spm}.
We can see that ChineseBERT generally outperforms MacBERT on LCQMC but slightly underperforms BERT-wwm. We hypothesis this is because the domain of BQ Corpus more fits the pretraining data of BERT-wwm than that of ChineseBERT.

\begin{table}[t]
  \centering
  \small
  \begin{tabular}{lcccc}
    \toprule 
     & \multicolumn{2}{c}{{\it LCQMC}} & \multicolumn{2}{c}{{\it BQ Corpus}}\\
    {\bf Model} & {\bf Dev} & {\bf Test} & {\bf Dev} & {\bf Test}\\
    \midrule
    & \multicolumn{4}{c}{\underline{\texttt{Base}}}\\
    ERNIE & 89.8 & 87.2 & 86.3 & 85.0 \\
    BERT & 89.4 & 87.0 & 86.1 & 85.2 \\
    BERT$^{\circ}$  & 89.6 & 87.1 & {\bf 86.4} & {\bf 85.3} \\
    RoBERTa$^{\circ}$  & 89.0 & 86.4 & 86.0 & 85.0 \\ 
    MacBERT  & 89.5 & 87.0 &  86.0 & 85.2 \\
    ChineseBERT  & {\bf 89.8} & {\bf 87.4} & {\bf 86.4} & 85.2 \\
    \cdashline{1-5}
    & \multicolumn{4}{c}{\underline{\texttt{Large}}}\\
    RoBERTa$^{\circ}$  & 90.4 & 87.0 & { 86.3} & { 85.8} \\
    MacBERT & {\bf 90.6} & 87.6 & 86.2 & 85.6 \\
    ChineseBERT  & 90.5 & {\bf 87.8} & {\bf 86.5} & {\bf 86.0} \\
    \bottomrule
  \end{tabular}
  \caption{Performances of different models on SPM datasets LCQMC and BQ Corpus. We report accuracy for comparison. $\circ$ represents models pretrained on extended data.}
  \label{tab:spm}
\end{table}

\subsection{Named Entity Recognition (NER)}
For NER tasks \cite{chiu2016named,lample2016neural,li2019unified},  
the model is asked to identify named entities within a piece of text, which is formalized as a sequence labeling task. We use OntoNotes 4.0 \citep{weischedel2011ontonotes} and Weibo \citep{peng2015named} for this task. 
We use OntoNotes 4.0 and Weibo NER for this task. 
OntoNotes has 18 named entity types and Weibo has 4 named entity types.  OntoNotes and Weibo respectively contain 15K/4K/4K and 1,350/270/270 instances for training/dev/test.
Results are shown in Table \ref{tab:ner}.
As we can see, ChineseBERT significantly outperforms BERT and RoBERTa in terms of F1. In spite of a slight loss on precision for the \texttt{base} version, the gains on recall are particularly high, leading to a final performance boost on F1.

\begin{table}[t]
  \centering
  \small
  \scalebox{0.8}{
  \begin{tabular}{lcccccc}
    \toprule 
     & \multicolumn{3}{c}{{\it OntoNotes 4.0}} & \multicolumn{3}{c}{{\it Weibo}}\\
    {\bf Model} & {\bf P} & {\bf R} & {\bf F} & {\bf P} & {\bf R} & {\bf F}\\\midrule
    & \multicolumn{5}{c}{\underline{\texttt{Base}}}\\
    BERT & 79.69&82.09&80.87&67.12&66.88&67.33 \\
    RoBERTa$^{\circ}$ & {\bf 80.43}& 80.30 &80.37&{\bf 68.49}&67.81&68.15\\
    ChineseBERT & 80.03&{\bf 83.33}&{\bf 81.65}&68.27& {\bf 69.78}&{\bf 69.02} \\
    \cdashline{1-7}
    & \multicolumn{5}{c}{\underline{\texttt{Large}}}\\
    RoBERTa$^{\circ}$ &80.72&82.07&81.39&66.74&70.02&68.35\\
    ChineseBERT &{\bf 80.77}&{\bf 83.65}&{\bf 82.18}&{\bf 68.75}&{\bf 72.97}&{\bf 70.80}\\
    \bottomrule
  \end{tabular}
  }
  \caption{Performances of different models on NER datasets OntoNotes 4.0 and Weibo.  Results of precision (P), recall (R) and F1 (F) on test sets are reported for comparison.}
  \label{tab:ner}
\end{table}

\subsection{Chinese Word Segmentation}
The task divides text into words and is formalized as a character-based sequence labelling task.
We use the PKU and MSRA datasets for Chinese word segmentation. 
PKU consists of 19K/2K sentences for training and test, and MSRA consists of 87k/4k sentences for training and test. Output character embedding is fed to the softmax function for final predictions. 
Results are shown in Table \ref{tab:cws}, where we can see that ChineseBERT is able to outperform BERT-wwm and RoBERTa-wwm on both datasets for both metrics.

\begin{table}[t]
  \centering
  \small
  \begin{tabular}{lcccc}
    \toprule 
     & \multicolumn{2}{c}{{\it MSRA}} & \multicolumn{2}{c}{{\it PKU}}\\
    {\bf Model} & {\bf F1} & {\bf Acc} & {\bf F1} & {\bf Acc}\\
    \midrule
    & \multicolumn{4}{c}{\underline{\texttt{Base}}}\\
    BERT$^{\circ}$  & 98.42 &99.04 & 96.82 &97.70 \\
    RoBERTa$^{\circ}$  &98.46 &99.10 & 96.88 &97.72 \\ 
    ChineseBERT  & {\bf 98.60}& {\bf 99.14}  &{\bf 97.02}& {\bf 97.81} \\
    \cdashline{1-5}
    & \multicolumn{4}{c}{\underline{\texttt{Large}}}\\
    RoBERTa$^{\circ}$  &98.49 &99.13 & 96.95& 97.80 \\
    ChineseBERT  &{\bf 98.67} &{\bf 99.26} &{\bf  97.16} &{\bf 98.01} \\
    \bottomrule
  \end{tabular}
  \caption{Performances of different models on CWS datasets MSRA and PKU. We report F1 and accuracy (Acc) for comparison. $\circ$ represents models pretrained on extended data.}
  \label{tab:cws}
\end{table}

\section{Ablation Studies}
In this section, we conduct ablation studies to understand the behaviors of ChineseBERT. 
We use the Chinese named entity  recognition dataset OntoNotes 4.0  for analysis and all models are based  on the \texttt{base} version.

\subsection{The Effect of Glyph Embeddings and Pinyin Embeddings}
We would like to explore the effects of glyph embeddings and pinyin embeddings.
For fair comparison,
we pretrained different models on the same dataset, with the same number of training steps, and with the same model size. 
Setups include ``-glyph'', where glyph embeddings are not considered and we only consider pinyin and char-ID embeddings;
``-pinyin'', where pinyin embeddings are not considered and we only consider glyph and char-ID embeddings; 
``-glyph-pinyin'', where only char-ID embeddings are considered, and the model degenerates to RoBERTa. 
We finetune different models on the OntoNotes dataset  of the NER dataset for comparison.  
Results are shown in Table \ref{tab:effect}. As can be seen, either removing glyph embeddings or pinyin embeddings results in performance degradation, and removing both has the greatest negative impact on the F1 value, which is a drop of about 2 points. This validates the importance of both pinyin and glyph embeddings for modeling Chinese semantics. 
The reason why ``-glyph-pinyin'' performs worse than RoBERTa is that 
the model we use here is trained  on a 
 smaller size of data with smaller number of training steps.

\begin{table}[t]
  \centering
  \small
  \begin{tabular}{lccl}
    \toprule 
     & \multicolumn{3}{c}{{\it OntoNotes 4.0}} \\
    {\bf Model} & {\bf Precision} & {\bf Recall} & {\bf F1}\\\midrule
   RoBERTa$^\circ$ & 80.43 &  80.30  &  80.37 \\
    ChineseBERT &  80.03   &  83.33 & 81.65\\
    -- Glyph & 77.67 & 82.75  & 80.13 (-1.52)\\
    -- Pinyin &77.54   & 83.65  &  80.48 (-1.17)\\
    -- Glyph -- Pinyin& 78.22  &  81.37  & 79.76 (-1.89)\\
    \bottomrule
  \end{tabular}
  \caption{Performances for different models without considering  glyph or pinyin information.}
  \label{tab:effect}
\end{table}

\subsection{The Effect of Training Data Size}
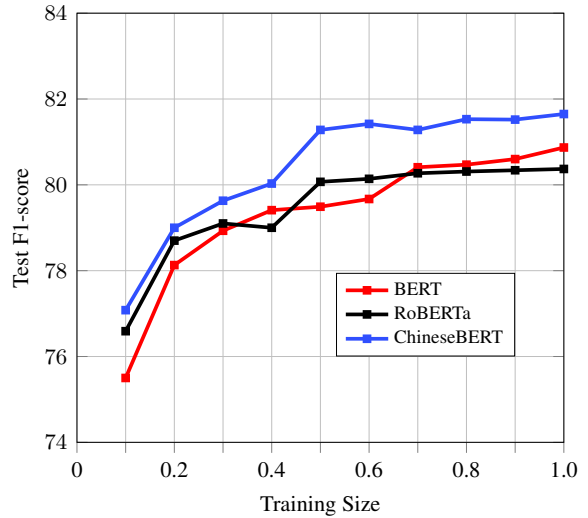
\begin{figure}[!ht]     
\centering
\begin{tikzpicture}[scale=0.93]
\begin{axis}[
    	   width=1.1\columnwidth,
	    height=1\columnwidth,
	    legend cell align=left,
	    legend style={at={(0.9, 0.2)},anchor=south east,font=\small,nodes={scale=0.85, transform shape}},
	    xtick={0,1,2,3,4,5,6,7,8,9,10},
	    xticklabels={0, , 0.2, , 0.4, , 0.6, , 0.8, ,1.0},
   		ytick={74, 76, 78, 80, 82, 84, 85},
        ymin=74, ymax=84,
   		xtick pos=left,
   		xtick align=outside,
	    xmin=0,xmax=10,
	    mark options={mark size=1},
		font=\small,
   	 	ymajorgrids=true,
    	xmajorgrids=true,
    	xlabel=Training Size,
        ylabel=Test F1-score,
    	ylabel style={at={(axis description cs: -0.08, 0.5)}}]
    	
\addplot[
    color=red,
    mark=square*,
    line width=1.5pt
    ]
    coordinates {
(1, 75.50)(2, 78.13)(3,78.93)(4,79.41)(5,79.49)(6,79.67) (7,80.41)(8,80.47)(9,80.60)(10,80.87)
    };
    \addlegendentry{BERT}

\addplot[
    color=black,
    mark=square*,
    line width=1.5pt
    ]
    coordinates {
(1, 76.59)(2, 78.7)(3,79.10)(4,79.00)(5,80.07)(6,80.14)(7,80.27)(8,80.31)(9,80.34)(10,80.37)
    };
    \addlegendentry{RoBERTa}

\addplot[
    color=shannon,
    mark=square*,
    line width=1.5pt
    ]
    coordinates {
(1, 77.08)(2, 79.0)(3,79.63)(4,80.03)(5,81.28)(6,81.42)(7,81.28)(8,81.53)(9, 81.52)(10, 81.65)
    };
    \addlegendentry{ChineseBERT}

\end{axis}
\end{tikzpicture}
\caption{Performances when varying the training size.}
\label{fig:size}
\end{figure}

We hypothesize glyph and pinyin embeddings also serve as strong regularization over text semantics, which means that the proposed ChineseBERT model is able to perform better with less training data.
We randomly sample 10\%$\sim$90\% of the training data while maintaining the ratio of samples with entities w.r.t. samples without entities. We perform each experiment five times and report the average F1 value on the test set.
Figure \ref{fig:size} shows the results. 
As can be seen, ChineseBERT performs better across all setups. 
With less than 30\% of the training data, the improvement of ChineseBERT is slight, but with over 30\% training data, the performance improvement is greater.
This is because ChineseBERT still requires sufficient training data to fully train the glyph and pinyin embeddings, and insufficient training data would lead to inadequate training. 

%

\section{Conclusion}
In this paper, we introduce ChineseBERT, a large-scale pretraining Chinese NLP model. It leverages the glyph and pinyin information of Chinese characters to enhance the model's ability of capturing context semantics from surface character forms and disambiguating polyphonic characters in Chinese.
 The proposed ChineseBERT model achieves significant performance boost across a wide range of Chinese NLP tasks.
The proposed ChineseBERT performs better than vanilla pretrained models with less training data, indicating that the introduced glyph embeddings and pinyin embeddings serve as a strong regularizer for semantic modeling in Chinese.
Future work involves training a large size version of ChineseBERT. 

\section*{Acknowledgement}
This work is supported by Key-Area Research and Development Program of Guangdong Province（No.2019B121204008). 
We also want to acknowledge 
 National Key R\&D Program of China (2020AAA0105200) and Beijing Academy of Artificial Intelligence (BAAI).

\end{CJK*}

\bibliographystyle{acl_natbib}
\bibliography{glyce}

\end{document}